\documentclass[runningheads]{llncs}
\usepackage[T1]{fontenc}
\usepackage{booktabs} 
\usepackage{pifont} 
\usepackage{graphicx}
\usepackage{amsmath}
\usepackage{booktabs}
\usepackage{tabularx} 
\usepackage{amssymb}
\usepackage{hyperref}
\usepackage{multirow}
\usepackage{graphicx}
\begin{document}

\title{SPOT-Face: Forensic Face Identification using Attention Guided Optimal Transport}

\titlerunning{SPOT-Face}
%
\author{Ravi Shankar Prasad\orcidID{0009-0003-7767-6762} \and
Dinesh Singh\orcidID{0000-0001-8889-9847} 
}
\authorrunning{R Prasad et al.}
\institute{
   Vision Intelligence and Machine Learning (VIML) Group, \\
  School of Computing and Electrical Engineering (SCEE), \\
  Indian Institute of Technology Mandi (IIT Mandi), Himachal Pradesh, India. \\
  \email{d23033@students.iitmandi.ac.in, dineshsingh@iitmandi.ac.in} \\
 }

\maketitle              
\begin{abstract}
Person identification in forensic investigations becomes very challenging when common identification means for DNA (i.e., hair strands, soft tissue) are not available. Current methods utilize deep learning methods for face recognition. However, these methods lack effective mechanisms to model cross-domain structural correspondence between two different forensic modalities. In this paper, we introduce a SPOT-Face, a superpixel graph-based framework designed for cross-domain forensic face identification of victims using their skeleton and sketch images. Our unified framework involves constructing a superpixel-based graph from an image and then using different graph neural networks(GNNs) backbones to extract the embeddings of these graphs, while cross-domain correspondence is established through attention-guided optimal transport mechanism. We have evaluated our proposed framework on two publicly available dataset: IIT\_Mandi\_S2F (S2F) and CUFS. Extensive experiments were conducted to evaluate our proposed framework. The experimental results show significant improvement in identification metrics ( i.e., Recall, mAP) over existing graph-based baselines. Furthermore, our framework demonstrates to be highly effective for matching skulls and sketches to faces in forensic investigations.
\keywords{Graph  \and  Superpixel \and Matching \and Retrieval\and Forensics.}
\end{abstract}
\section{Introduction}
In forensic investigations, identifying an unknown deceased individual is the first task forensic team tries to attempt. However, in case of skull identification, this task becomes challenging when conventional biometric evidences (i.e, fingerprints, soft tissue) are not available. Additionally, when someone wants to conduct DNA testing on a deceased individual, a key question arises about whom the DNA should be matched against. Therefore, to narrow down the possibilities for matching and to expedite the process, skull-based identification plays a crucial role. Both skull and sketch has a unique structure that corresponds directly to an individual's face and the study done in~\cite{duan2014skull} states that there is one-to-one mapping between the skull and its corresponding facial appearance. Thus, our aim is to learn a common representation that can be used for cross-domain matching. For skull-to-face identification, forensic artists create a reconstruction of the deceased's face by applying a layer of clay to the skull, guided by the thickness of the soft tissue information~\cite{hona2024global}, whereas for sketch-to-face matching artists draw sketches images based on the description of victims or witnesses. However, forensic artists often don't have full details about a person's facial features, like the shape of their nose, ears, or eye color. Instead, they use their creativity and experience to piece together a representation of the face. Due to these factors, skull and sketch images have significant modality discrepancies when compared to facial optimal images. As a result, making standard face recognition systems~\cite{li2020review} ineffective for direct matching. The algorithm~\cite{tang2004face} designed for sketch-to-face recogniton purpose generates a pseudo-photo or sketch by utilizing the Karhunen-Loeve Transform. Following this, a nonlinear dimension reduction method~\cite{liu2005nonlinear} was proposed to produce more realistic sketch photos, aiming to increase recognition accuracy. While these methods improved visual similarity but they struggled to generalize to complex forensic scenarios such as large modality gaps, limited training data.

Skull-face recognition is more challenging than sketch-face due to the large modality gap between skull and face images compared to sketch and face as shown in Figure~\ref{fig:tsne}. To address this challenge, we propose a superpixel~\cite{achanta2012slic} graph-based framework which explicitly learns cross-domain correspondences using attention-guided optimal transport. By capturing both local structural relationships and global cross-modal alignment, the proposed approach enables effective skull-to-face and sketch-to-face matching in a unified representation space.
The contributions of our paper are as follows:

\begin{itemize}
    \item  We introduce a novel SPOT-Face graph oriented cross-attention and optimal transport based framework for cross-domain identificaion from skulls and sketches.
    \item We evaluated our framework with different Graph Neural Networks (GNN) backbones with different cross modules validation studies. 
    \item We have conducted extensive and comprehensive evaluation for skull-to-face and sketch-to-face matching using different metrics such as recall@k and mAP@K on two public available dataset IIT\_Mandi\_S2F~\cite{prasad2025fcr} and CUFS~\cite{zhang2011coupled}.
\end{itemize}

 \begin{figure*}[!ht]
        \centering
        \includegraphics[width =\textwidth,keepaspectratio,trim={2.5cm 2.7cm 2cm 1.0cm},clip]{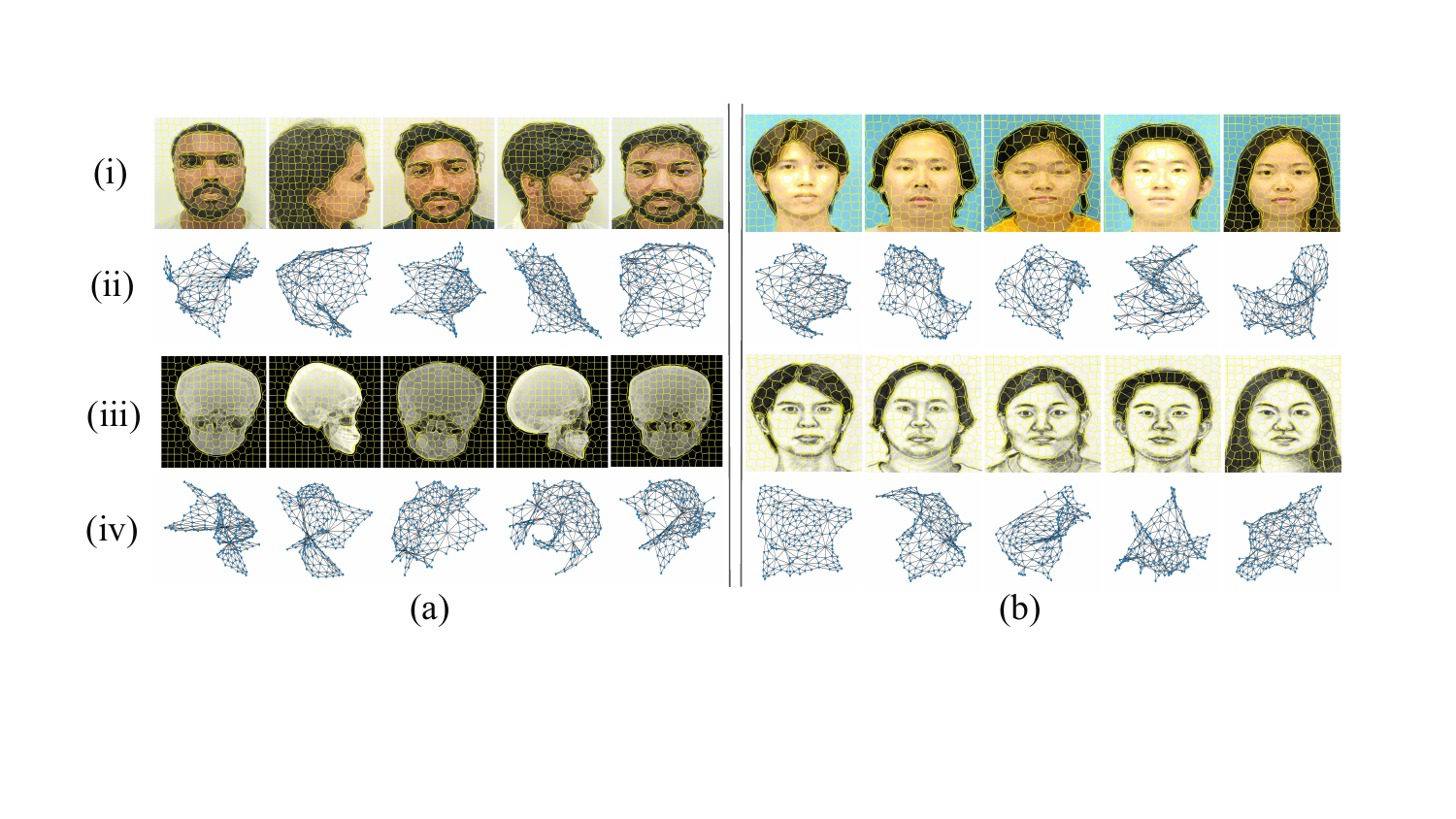}
        \caption{(a) shows paired face and their X-ray samples from IIT\_Mandi\_S2F dataset where (b) represents the paired face and their sketch samples from CUFS dataset with superpixel representation and their graph.}
        \label{fig:dataset}
    \end{figure*}  
\section{Related Work}
\subsection{Skull-to-face}
In forensic investigations, person's identification through image retrieval aims to identify the correct persons for the given query images ( i.e, skull) from the large gallery of face dataset. Few works has been done on skull-face recognition using superimposition~\cite{damas2011forensic,damas2020meprocs} and reconstruction~\cite{claes2010bayesian,claes2010computerized} methods. But, all these methods uses approximate analysis for craniofacial identifications. Also, work done by~\cite{prasad2025cross,prasad2025fcr} present deep learning based cross-domain (skull-face) alignment using contrastive and adversarial loss, however these methods are more burdened with the model capabilities. Hence, to tackle of these challenges, we have introduced cross-attention and optimal-transport mechanism to enhance and align features of cross-domain modalities.
\subsection{Sketch-to-face}
Tradionally, plenty of work done by~\cite{klare2011matching,galoogahi2012intermodality,alex2013ldgbp} on forensic sketch-face recognition, however these methods rely on manually designed descriptors with limited representation capacity and poor generalization. With the advacements in deep learning methods, several studies are conducted for sketch-face image recognition and retrieval such as~\cite{misra2017red,cakir2019deep,roth2019mic,sanakoyeu2019divide,wang2020cross}.

The work done in~\cite{fan2021siamesegcn} proposed the use of Siamese graph GCNs network to match the sketch with its corresponding face image, but this work mainly focuses on training deep models with contrastive loss with no cross-domain embeddings alignment. Later for features alignment, work proposed in~\cite{chen2022sketchtransformer,sain2021stylemeup} conducted their studies in features disentanglement for better features extraction in cross-domain modalities. GAN-based methods like Identity-Aware CycleGAN~\cite{fang2020identity} create photo-like images from sketches and then recognize these generated images. While these methods help keep the identity of people clear in the final images, they rely on an extra step in generating the images and do not allow for directly retrieving faces from sketches. In contrast, our method matches sketches and skulls directly to face images in a single space and uses standard metrics, like Recall@k and mean Average Precision (mAP@k), to evaluate its performance.

Although above methods achieve promising performance in sketch-to-face recognition and retrieval but fails to capture the semantic correspondence between two different modalities. Hence, our proposed method bridges the gap between these cross-domain modalities using graph based features with leveraging cross-attention and optimal-transport mechanisms to find the meaningful semantic relationship between two modalities from different domains.

\begin{figure}[!ht]
        \centering
         \includegraphics[width=\linewidth,keepaspectratio,trim={0cm 0cm 0cm 0cm},clip]{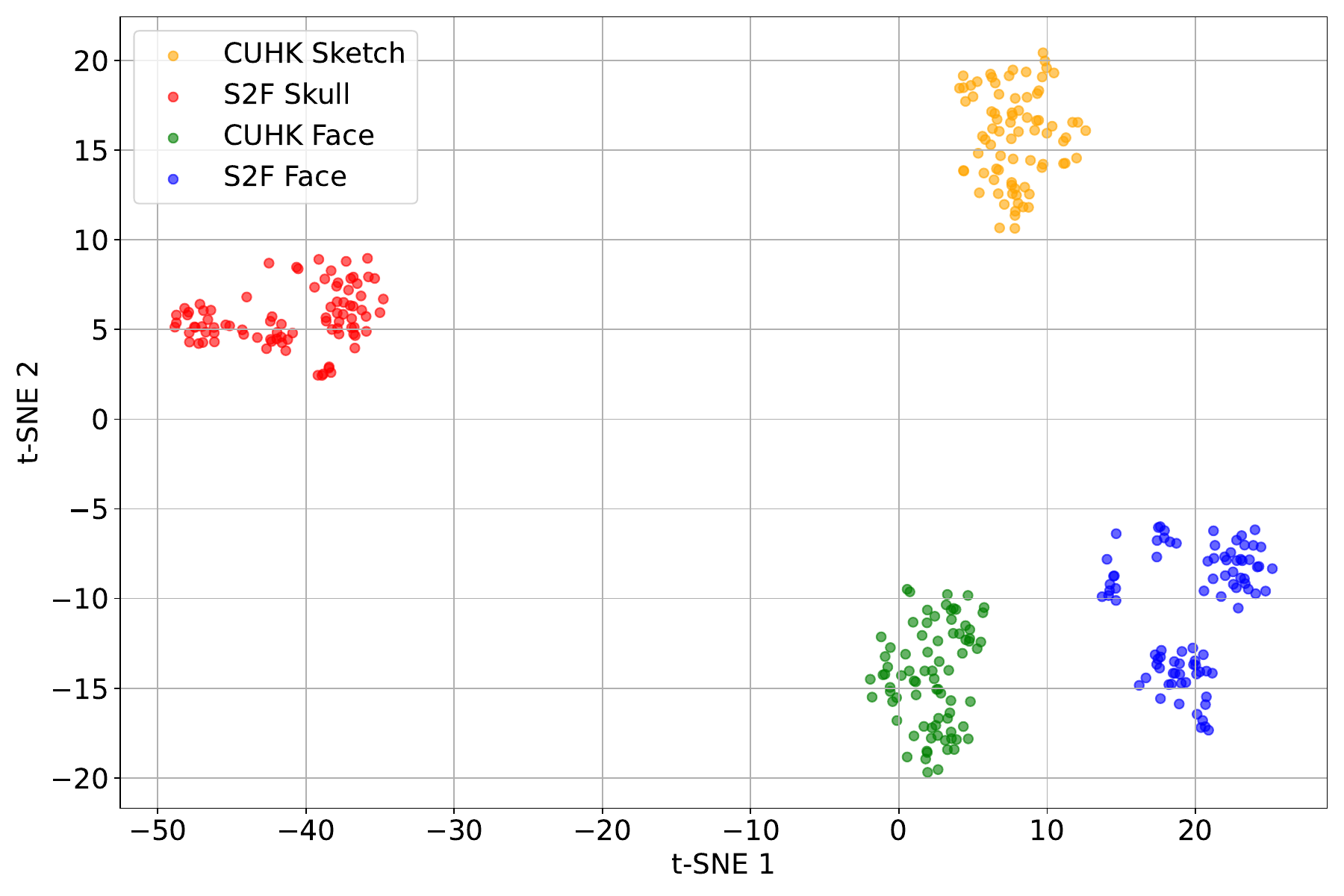}
        \caption{Visualization of embeddings of S2F and CUFS dataset. This 2D t-SNE shows that S2F (skull-to-face) dataset have larger domain gap than CUFS (sketch-to-face) dataset. (Best viewed in colors) }
        \label{fig:tsne}
    \end{figure}
\section{Dataset}\label{dataset_s}
 \paragraph{IIT\_Mandi\_S2F (S2F).} This dataset consists of 51 Indian paired face-skull~\cite{prasad2025fcr} images. In this dataset, face and skull images have both frontal and lateral view 2D images. Hence, we have total 102 images for face and 102 for its corresponding skull, contributing a total of 102 pairs of skull-face images. Skull images are the X-ray images and face images are the optimal images.

\paragraph{CUFS.} This dataset consists of paired 188 Chinese face-sketch~\cite{zhang2011coupled} images also known as CUHK face sketch dataset. This dataset contains only frontal pose face-sketch images. Figure~\ref{fig:dataset} shows some sample of images of S2F and CUFS dataset with their superpixel and graph representations.

\section{Methodology}
\begin{figure*}[!ht]
        \centering
        \includegraphics[width=\textwidth,keepaspectratio,trim={0.cm 0.25cm 0.5cm 0.0cm},clip]{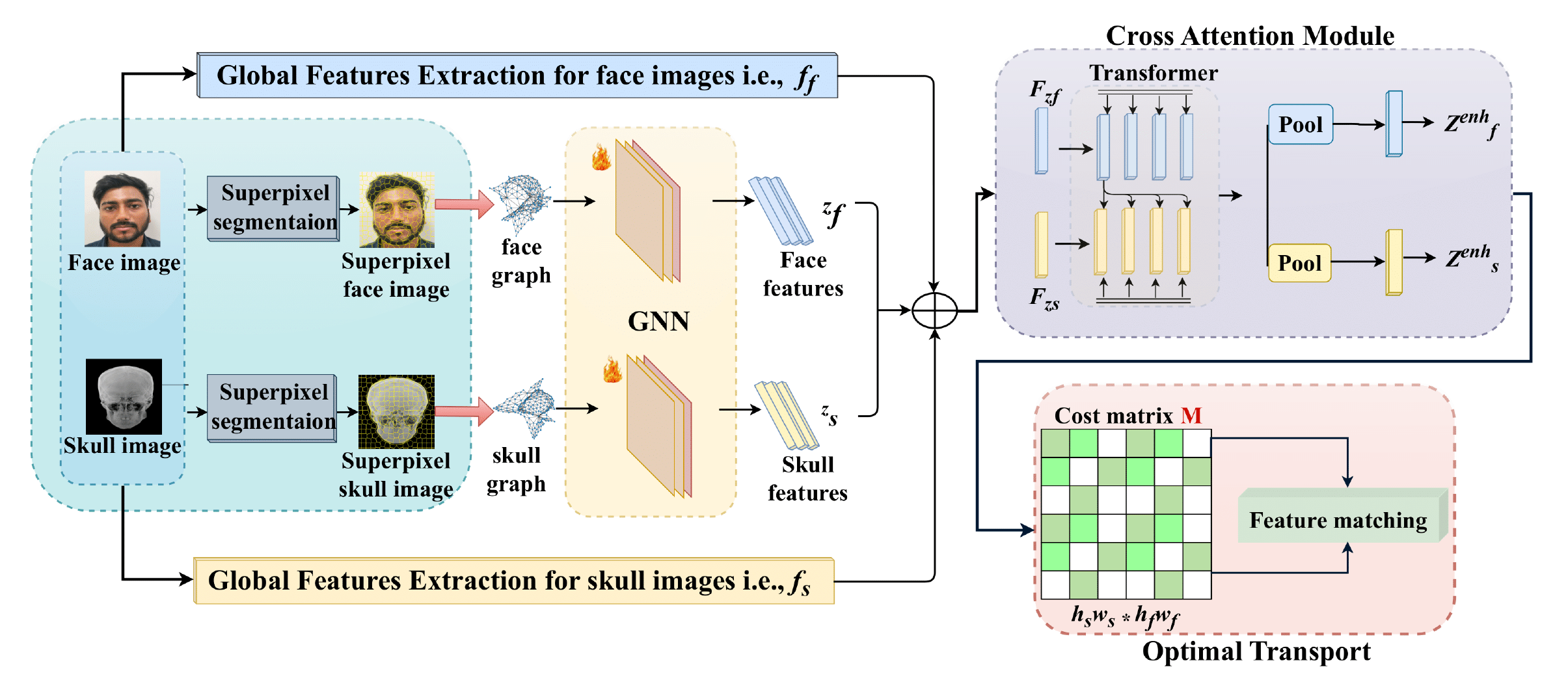}
        \caption{In this framework, different GNNs serves as the backbone to extract features from the input skull-face and sketch-face graphs. First, skull, sketch and their respective face images are converted to superpixels using the SLIC method, and from superpixel segmentation of images, we get the graphs by treating each segment as a node. Then, this graph is input to the trainable GNN module. Finally, to enhance the features of the skull, sketch and face graphs, we leverage a cross-attention and optimal transport module. We perform feature matching by minimizing the triplet loss. (Best viewed in colors)}
        \label{fig:framework}
    \end{figure*}

This paper introduces a new method that uses graph-structured data to reduce the cross-domain gap between face-skull in S2F and face-sketch in CUFS dataset for face recognition system. We apply a superpixel method to built graph from the images and then different graph neural networks are used to extract the features of respective graph. To get the best match between corresponding graph features, we utilize cross-attention and optimal transport module as shown in Figure~\ref{fig:framework}.

\subsection{Graph Structure Data for Images}
Faces are created using optical imaging principles. A facial photograph captures the relationships between pixels to depict all the features of a real human face in a two-dimensional space, and for skull and sketch images, there is no such relationship between pixels to represent all the features of a human skull because a skull has only a bone-like structure and the sketch has only boundary based structural information. This results in a significant gap between these two modalities. As the representations of skull-sketch and their respective face differ, convolutional networks can only capture the local structures of the skull and face, but cannot fully extract color and texture information. As a result, they fail to represent the same model features as the skull or sketch, since the nature of the skull/sketch leads to a lack of “feature sense” in the extracted data.

To address this issue, we propose building a graph structure based on the features and structural information of the images, as illustrated in Figure~\ref {fig:framework}. In the first step, we employed the methods as mentioned in~\cite{xie2015holistically,fan2021siamese} for edge detection method and to extract the contours of the images (i.e, skull, sketch, face) and to minimize background noise. This approach preserves the structural properties of the image while reducing irrelevant information. Input images are then converted to graph-structured data using superpixel methods~\cite{vedaldi2008quick,achanta2012slic}. It uses the center point of each region as a node in a graph; hence, the total number of nodes in the graph is equal to the number of segmented regions. The edges of the graph show the distance between each pair of nodes. This organized structure helps simplify further analysis and processing of the image by focusing on the relationships between the segmented regions.

\subsection{\textbf{Learning Cross-Domain Identity Representation}}
We employed four GNNs with cross-attention and optimal transport module to achieve cross-domain identity representation by calculating the matching score between the provided skull and face images. We incorporated different graph neural networks (i.e, GCN, GAT, Graphsage, GraphTransformer) models as the backbone to extract embeddings from both face, skull and sketch images. Since skull images originate from a different domain, we maintained the same backbone architecture but added learnable layers to train the embeddings specifically for the skull images. We have made the channel trainable for the both input skull and face graph in S2F dataset, and also for the CUFS dataset, as shown in Figure~\ref{fig:framework}. Finally, we leverage cross-attention to get more enhanced features from respective graph features and optimal-transport module to get the optimal match.

\subsection{Graph-Based Cross-Modal Alignment}

Let $\mathcal{I}_s$, $\mathcal{I}_k$, and $\mathcal{I}_f$ denote skull, sketch, and face images, respectively. 
Each image is first segmented into $N$ superpixels using the SLIC algorithm~\cite{achanta2012slic}. 
From the superpixel segmentation, a graph 
$\mathcal{G}=(\mathcal{V},\mathcal{E})$ is constructed, where each node 
$v_i \in \mathcal{V}$ corresponds to a superpixel and edges 
$e_{ij} \in \mathcal{E}$ connect spatially adjacent superpixels.

For each modality $m \in \{s,k,f\}$, we obtain an input graph 
$\mathcal{G}_m=(\mathcal{V}_m,\mathcal{E}_m)$ with initial node features 
$\mathbf{X}_m \in \mathbb{R}^{N_m \times d_0}$, then different GNN encoders $f_m(\cdot)$ are employed to extract node embeddings which is given as:
\begin{equation}
\mathbf{H}_m = f_m(\mathcal{G}_m; \theta_m), \qquad m \in \{s,k,f\},
\end{equation}
where $\mathbf{H}_m \in \mathbb{R}^{N_m \times d}$ denotes the learned node-level features.

To enhance cross-modal correspondence, we introduce a feature refinement module composed of cross-attention (CA) and optimal transport (OT). 
Given a pair of modalities $(m,n)$, cross-attention is defined as:
\begin{equation}
\tilde{\mathbf{H}}_m =
\mathrm{softmax}\!\left(
\frac{\mathbf{H}_m \mathbf{W}_Q (\mathbf{H}_n \mathbf{W}_K)^\top}{\sqrt{d}}
\right)
\mathbf{H}_n \mathbf{W}_V,
\end{equation}
where $\mathbf{W}_Q, \mathbf{W}_K, \mathbf{W}_V \in \mathbb{R}^{d \times d}$ are learnable parameters, then, the residual connections and normalization yield enhanced features are given as:
\begin{equation}
\hat{\mathbf{H}}_m = \mathrm{LN}\!\left(\mathbf{H}_m + \tilde{\mathbf{H}}_m \right).
\end{equation}

To further enforce local alignment, we formulate an entropic OT problem between enhanced node embeddings $\hat{\mathbf{H}}_m$ and $\hat{\mathbf{H}}_n$. 
The cost matrix $\mathbf{C}_{mn} \in \mathbb{R}^{N_m \times N_n}$ is defined using cosine distance:
\begin{equation}
\mathbf{C}_{mn}(i,j) =
1 - \frac{\hat{\mathbf{h}}_{m,i}^\top \hat{\mathbf{h}}_{n,j}}
{\|\hat{\mathbf{h}}_{m,i}\|_2 \|\hat{\mathbf{h}}_{n,j}\|_2}.
\end{equation}

The optimal transport plan is obtained as:
\begin{equation}
\mathbf{T}^* =
\arg\min_{\mathbf{T} \in \Pi(\mu,\nu)}
\langle \mathbf{T}, \mathbf{C}_{mn} \rangle
- \varepsilon H(\mathbf{T}),
\end{equation}
where $\Pi(\mu,\nu)$ denotes the transport polytope with prescribed marginals, and 
$H(\mathbf{T}) = -\sum_{i,j} \mathbf{T}_{ij} \log \mathbf{T}_{ij}$ is the entropy regularizer. After this, global graph-level embeddings are obtained by mean pooling:
\begin{equation}
\mathbf{z}_m =
\mathrm{Norm}\!\left(
\frac{1}{N_m} \sum_{i=1}^{N_m} \hat{\mathbf{h}}_{m,i}
\right).
\end{equation}

During training, feature matching is enforced using a triplet loss. 
Given an anchor graph embedding $\mathbf{z}_a$, a positive $\mathbf{z}_p$, a negative $\mathbf{z}_n$ and $\alpha$ as the margin, the loss is defined as:
\begin{equation}\label{triplet_loss}
\mathcal{L}_{\text{tri}} =
\max\!\left(0,
\|\mathbf{z}_a - \mathbf{z}_p\|_2^2
- \|\mathbf{z}_a - \mathbf{z}_n\|_2^2 + \alpha
\right).
\end{equation}

This objective encourages matched skull--face and sketch--face pairs to be closer than mismatched pairs, while the cross-attention and OT modules promote cross-modal feature alignment.

\subsection{Image retrieval framework}
A retrieval framework is essential in forensic applications because it allows for the efficient and accurate identification of unknown individuals through similarity-based searches in large databases. There were several studies conducted for image retrieval in many different application, as mentioned in~\cite{zaeemzadeh2021face}. When provided with a query image (skull/sketch), the system retrieves the most relevant images (face) from a gallery.

In forensic applications, recall, mAP ( mean average precision) are important retrieval metrics~\cite{manning2009introduction}. Recall ensures that true identities are not missed, even if they do not appear as the top match while mAP takes into account both the ranking of the results and the presence of relevant items, making it as a comprehensive metric. In forensic contexts, where multiple relevant images or identities may need to be retrieved, recall@k and mAP@k effectively captures how well the system ranks all potential matches. 


\begin{table}[t]
\centering
\caption{
Impact of graph KNN hyperparameter ($k$) on Skull--Face (IIT\_Mandi\_S2F) and
Sketch--Face (CUFS) datasets. Here,
Recall@K and mAP@K (\%) are reported and the best results are shown in bold.}
\label{tab:k_ablation_combined}

\setlength{\tabcolsep}{6pt}
\renewcommand{\arraystretch}{1.1}
\small

\begin{tabular}{c|cccc|cccc}
\hline
\multirow{2}{*}{\textbf{$k$}} &
\multicolumn{4}{c|}{\textbf{Skull--Face (S2F)}} &
\multicolumn{4}{c}{\textbf{Sketch--Face (CUFS)}} \\
\cline{2-9}

 & R@1 & R@5 & mAP@1 & mAP@5 &
   R@1 & R@5 & mAP@1 & mAP@5 \\
\hline

4  & 50.0 & 71.0 & 50.0 & 64.5 &
     88.4 & 89.0 & 88.4 & 88.7 \\

6  & \textbf{50.0} & \textbf{72.0} & \textbf{50.0} & \textbf{65.4} &
     88.4 & 88.9 & 88.4 & 88.7 \\

8  & 49.0 & 66.1 & 49.0 & 60.6 &
     88.4 & 89.0 & 88.4 & 88.8 \\

12 & 50.0 & 72.0 & 50.0 & 64.1 &
     \textbf{88.4} & \textbf{89.0} & \textbf{88.4} & \textbf{88.9} \\
\hline
\end{tabular}
\end{table}

\begin{table}[t]
\centering
\caption{
Comparison of different GNN backbones on Skull--Face (IIT\_Mandi\_S2F)
and Sketch--Face (CUFS) datasets. Here,
Recall@K and mAP@K (\%) are reported and the best results are shown in bold.}
\label{tab:gnn_backbone_sequential}

\setlength{\tabcolsep}{6pt}
\renewcommand{\arraystretch}{1.1}
\small

\begin{tabular}{lcccc}
\hline
\multicolumn{5}{c}{\textbf{Skull--Face (IIT\_Mandi\_S2F)}} \\
\hline
\textbf{Backbone} & \textbf{R@1} & \textbf{R@5} & \textbf{mAP@1} & \textbf{mAP@5} \\
\hline
GCN              & 50.0 & 71.0 & 50.0 & 63.9 \\
GAT              & 49.5 & 68.1 & 49.5 & 62.0 \\
GraphSAGE        & 50.0 & 70.1 & 50.0 & 63.4 \\
GraphTransformer &
\textbf{50.0} & \textbf{72.0} & \textbf{50.0} & \textbf{65.4} \\
\hline
\hline

\multicolumn{5}{c}{\textbf{Sketch--Face (CUFS)}} \\
\hline
\textbf{Backbone} & \textbf{R@1} & \textbf{R@5} & \textbf{mAP@1} & \textbf{mAP@5} \\
\hline
GCN (Quickshift)~\cite{fan2021siamese} & 82.4 & -- & -- & -- \\
GCN (SLIC)~\cite{fan2021siamese}       & 87.7 & -- & -- & -- \\
\hline
GCN              & 88.4 & 88.9 & 88.4 & 88.8 \\
GAT              & 88.4 & 88.6 & 88.4 & 88.6 \\
GraphSAGE        & 88.4 & 88.8 & 88.4 & 88.6 \\
GraphTransformer &
\textbf{88.4} & \textbf{89.0} & \textbf{88.4} & \textbf{88.9} \\
\hline
\end{tabular}
\end{table}
\begin{figure}[t]
        \centering
         \includegraphics[width=\linewidth,keepaspectratio,trim={0.3cm 2.cm 0cm 1cm},clip]{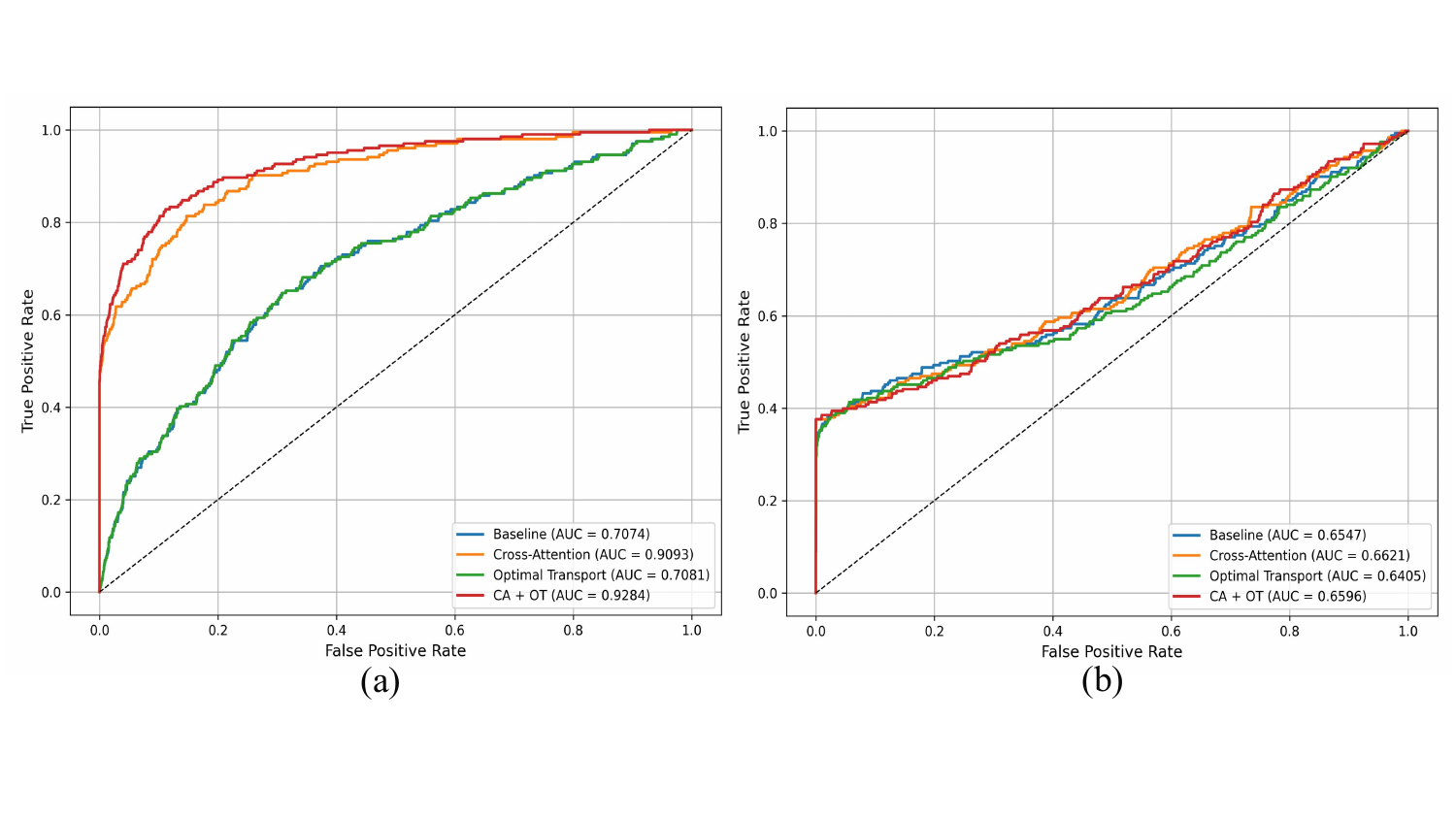}
        \caption{(a) and (b) show ROC\_AUC curve for IIT\_Mandi\_S2F and CUFS dataset with different modules, respectively. (Best viewed in colors)}
        \label{fig:qualit1}
    \end{figure}

\newcommand{\cmark}{\ding{51}} 
\newcommand{\xmark}{\ding{55}} 


\section{Experiments}
A comprehensive validation of the automatic skull-sketch to face matching system is essential for its practical application in real-life scenarios. It provides scientific evidence to support the increased utility of this recognition technique in criminal investigations. The proposed framework for skull-face and sketch-face similarity matching is evaluated using S2F and CUFS dataset. We have evaluated our model with loss functions as mentioned in \emph{Eq.}~\ref{triplet_loss}.

 \textbf{Implementation details.} The proposed method is implemented using a 24GB Nvidia RTX A500 GPU graphics card, designed for complex deep learning computations. We trained our proposed framework, which uses the CA-OT model, for 50 epochs. We selected the best hyperparameters: number of nodes (i.e., segments) in graph are 300 for both S2F and CUFS dataset, a learning rate of 0.0001, a batch size of 16, a margin of 0.3 for the triplet loss, a weight decay of 0.00001, 80 sinkhorn iterations, and 4 attention heads. For training our proposed model with S2F dataset, we partitioned this dataset into training, validation and test with the split ratio of 70:20:10 respectively. We have also maintained the same split ratio for CUFS dataset. We kept the same number of training iterations for both the S2F and CUFS datasets.


\begin{figure}[!ht]
        \centering
        \includegraphics[width=\textwidth, keepaspectratio,trim={6cm 4cm 7.5cm 2.cm},clip]{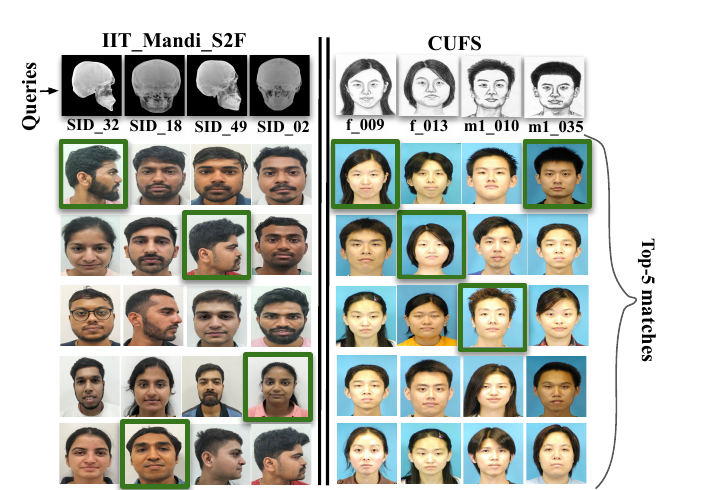}
        \caption{Qualitative evaluation for face retrieval showing top-5 retrieved faces. Where the green border face is the true face for the given query skull and sketch in IIT\_Mandi\_S2F and CUFS dataset. (Best viewed in colors)}
        \label{fig:qualit3}
    \end{figure}

\section{Results}
Quantitative and qualitative results are mentioned in section~\ref{quantitative} and~\ref{qualitative} for both S2F and CUFS dataset.

\begin{table}[t]
\centering
\caption{
Ablation of Cross-Attention (CA) and Optimal-Transport (OT) modules
on Skull–Face (S2F) and Sketch–Face (CUFS) matching.
Recall@K and mAP@K (\%) are reported.
Best results are shown in bold.}
\label{tab:ablation_ca_ot_combined}

\setlength{\tabcolsep}{5pt}
\renewcommand{\arraystretch}{1.1}
\small

\begin{tabular}{cc|cccc|cccc}
\hline
\multirow{2}{*}{\centering\textbf{CA}} &
\multirow{2}{*}{\centering\textbf{OT}} &
\multicolumn{4}{c|}{\textbf{Skull--Face (S2F)}} &
\multicolumn{4}{c}{\textbf{Sketch--Face (CUFS)}} \\
\cline{3-10}

 & &
R@1 & R@5 & mAP@1 & mAP@5 &
R@1 & R@5 & mAP@1 & mAP@5 \\
\hline

\xmark & \xmark &
31.3 & 46.0 & 31.3 & 38.1 &
83.4 & 88.7 & 83.4 & 85.4 \\

\cmark & \xmark &
48.5 & 60.7 & 48.5 & 56.8 &
88.4 & 88.8 & 88.4 & 88.7 \\

\xmark & \cmark &
31.3 & 45.5 & 31.3 & 37.7 &
83.4 & 88.8 & 83.4 & 85.8 \\

\cmark & \cmark &
\textbf{50.0} & \textbf{72.0} & \textbf{50.0} & \textbf{65.4} &
\textbf{88.4} & \textbf{89.0} & \textbf{88.4} & \textbf{88.9} \\

\hline
\end{tabular}
\end{table}

\subsection{Quantitative Results:}\label{quantitative}
Table~\ref{tab:k_ablation_combined} shows how number of neighbour (k) impact the image retrieval metrics recall@k and mAP@k. For S2F dataset, k = 6 achieves 50(\%), 72(\%) and 50(\%), 65.4(\%) for R@1, R@5 and mAP@1, mAP@5 as the best KNN hyperparatmer whereas for CUFS dataset, k = 12 achieves 88.4(\%), 89(\%) and 88.4(\%), 88.9(\%) for R@1, R@5 and mAP@1, mAP@5 in sketch-face recognition. These two table also show that retrieving face image for given query skull image in S2F dataset is more challenging than CUFS face-sketch dataset due to larger domain gap in S2F dataset as compared to CUFS dataset.

\subsection{Qualitative Results:}\label{qualitative}
Figure~\ref{fig:tsne} shows the embeddings distribution in S2F and CUFS dataset respectively. This 2D t-SNE shows that skull-face in S2F dataset have larger domain gap as compared to that of sketch-face in CUFS dataset. The reason behind larger domain gap in skull-face is due to lack of facial attributes (i.e, color, texture) in skull image. Figure~\ref{fig:qualit1} shows ROC-AUC curve for S2F and CUFS dataset with different module configuration. The model demonstrates strong verification capability, achieving a global ROC-AUC of 0.92 with S2F dataset, indicating effective separation between genuine and impostor pairs. Although the global ROC-AUC is 0.65 for CUFS dataset, this is because, the model is optimized for retrieval rather than verification, emphasizing relative ranking over absolute score separation. Figure~\ref{fig:qualit3} represents top-5 face retrieved for given query skull and sketch images. Hence, combined cross-attention and optimal-transport module achieves best performance in both dataset.

\section{Ablation Studies}
We evaluated our proposed framework different GNNs backbones for feature extraction and with different module configurations. Table~\ref{tab:gnn_backbone_sequential} shows how different graph neural networks ( GCN, GAT GraphSAGE, GraphTransformer) performs in extracting features from the respective graphs. Out of four backbone GraphTransformer is best in extracting features, achieving 50(\%), 72(\%) and 50(\%), 65.4(\%) for R@1, R@5 and mAP@1, mAP@5 in S2F dataset and 88.4(\%), 89(\%) and 88.4(\%), 88.9(\%) for R@1, R@5 and mAP@1, mAP@5 in CUFS dataset. To test the effectiveness of each module, we have conducted comparative studies with results in Table~\ref{tab:ablation_ca_ot_combined} on different module configurations. Also, our proposed method outperformed the work by Zhang et al.~\cite{fan2021siamesegcn} in sketch-face recognition achieving 88.4(\%), 89(\%) and 88.4(\%), 88.9(\%) for R@1, R@5 and mAP@1, mAP@5 on CUFS dataset.

\section{Conclusions}

In this paper, we present a SPOT-Face framework for skull-face and sketch-face recognition and retrieval using graph-structured data. Our proposed model incorporates cross-attention and optimal-transport module to improve the cross-domain alignment. Experimental results show that GraphTransformer serves better as a backbone in extracting graph features and our framework achieves better top-1 and top-5 recall and mAP than the others methods on the CUFS dataset. Hence, our proposed framework can be used as an efficient tool in forensics.

\end{document}